%% file: main.tex
\documentclass[lettersize,journal]{IEEEtran}
\usepackage{amsmath,amsfonts}
\usepackage{algorithmic}
\usepackage{algorithm}
\usepackage{array}
\usepackage[caption=false,font=normalsize,labelfont=sf,textfont=sf]{subfig}
\usepackage{textcomp}
\usepackage{stfloats}
\usepackage{url}
\usepackage{verbatim}
\usepackage{graphicx}
\usepackage{cite}
\usepackage{color}
\definecolor{myblue}{rgb}{0.16,0.44,0.7}
\usepackage{makecell}
\begin{document}

\title{Leveraging Reward Consistency for Interpretable Feature Discovery in Reinforcement Learning}

\author{Qisen Yang, Huanqian Wang, Mukun Tong, Wenjie Shi, Gao Huang,~\IEEEmembership{Member,~IEEE}, \\and Shiji Song,~\IEEEmembership{Senior Member,~IEEE}
\thanks{This work was supported in part by the Key-Area Research and Development Program of Guangdong Province under Grant 2020B1111500002, in part by the National Natural Science Foundation of China under Grants 62022048 and 62276150, and THU-Bosch JCML. \textit{(Corresponding author: Shiji Song.)}}
\thanks{Qisen Yang, Huanqian Wang, Mukun Tong, Gao Huang, and Shiji Song are with the Department of Automation, Tsinghua University, Beijing 100084, China. Email: \{yangqs19, wang-hq23, tmk19\}@mails.tsinghua.edu.cn, \{gaohuang, shijis\}@tsinghua.edu.cn. Wenjie Shi is with Meituan, Beijing 100102, China. Email: shiwenjie05@meituan.com.}}

\markboth{IEEE TRANSACTIONS ON SYSTEMS, MAN, AND CYBERNETICS: SYSTEMS}%
{Shell \MakeLowercase{\textit{et al.}}: A Sample Article Using IEEEtran.cls for IEEE Journals}

\IEEEpubid{\begin{minipage}{\textwidth}\ \\[20pt]
\fbox{
\parbox{0.975\textwidth}{
\copyright 2023 IEEE. Personal use of this material is permitted. Permission from IEEE must be obtained for all other uses, in any current or future media, including reprinting/republishing this material for advertising or promotional purposes, creating new collective works, for resale or redistribution to servers or lists, or reuse of any copyrighted component of this work in other works. 
}}\end{minipage}}

\maketitle

\begin{abstract}
    \input{chapters/0_abstract}
\end{abstract}


\section{Introduction}
    \input{chapters/1_introduction}

\section{Related Work}
    \input{chapters/2_related_work}

\section{Preliminaries}
    \input{chapters/3_preliminaries}

\section{Method}
    \input{chapters/4_method}

\section{Validity of Our Method}
\label{sec: Validity of Our Method}
    \input{chapters/5_1experiments}

\section{Understanding reward and action matching}
\label{sec: Understanding reward and action matching}
    \input{chapters/5_2experiments}

\section{Conclusion}
    \input{chapters/7_conclusion}

\appendices
\section{Experimental Details}
\label{appendixA}
\input{chapters/8_appendixA}
\section{Further Studies}
\label{appendixB}
\input{chapters/8_appendixB}

\bibliographystyle{IEEEtran}
\small{
    \bibliography{reference/ref.bib}
}

\end{document}

%% file: chapters/0_abstract.tex
The black-box nature of deep reinforcement learning (RL) hinders them from real-world applications. 
Therefore, interpreting and explaining RL agents have been active research topics in recent years.
Existing methods for post-hoc explanations usually adopt the action matching principle to enable an easy understanding of vision-based RL agents.
In this paper, it is argued that the commonly used action matching principle is more like an explanation of deep neural networks (DNNs) than the interpretation of RL agents.
It may lead to irrelevant or misplaced feature attribution when different DNNs' outputs lead to the same rewards or different rewards result from the same outputs. 
Therefore, we propose to consider rewards, the essential objective of RL agents, as the essential objective of interpreting RL agents as well.
To ensure reward consistency during interpretable feature discovery, a novel framework (RL interpreting RL, denoted as RL-in-RL) is proposed to solve the gradient disconnection from actions to rewards. 
We verify and evaluate our method on the Atari 2600 games as well as Duckietown, a challenging self-driving car simulator environment.
The results show that our method manages to keep reward (or return) consistency and achieves high-quality feature attribution.
Further, a series of analytical experiments validate our assumption of the action matching principle's limitations. 

%% file: chapters/1_introduction.tex
\IEEEPARstart{R}{einforcement} learning (RL) learns to solve sequential decision-making problems in an interactive environment. Combined with the powerful approximation capability of deep neural networks (DNNs), deep RL has soared and achieved impressive successes in various fields such as video games \cite{vinyals2019grandmaster} and robotics \cite{akkaya2019solving,DBLP:journals/tsmc/LiDGLLD20}. However, the black box nature of DNNs makes deep RL harder to understand, while trust and reliability are critical concerns in real-world applications, especially for high-stake scenarios like autonomous driving~\cite{DBLP:journals/tsmc/XuZLQRS20,DBLP:journals/tsmc/HanNFRO22} and medical care \cite{glanois2021survey}. Hence, exploring the underlying decision-making process and its interpretation (or explanation)~\footnote{Since no consensus about the nomenclature has been reached yet~\cite{glanois2021survey}, we use interpretation and explanation as synonyms in this paper.} will greatly contribute to more applicable RL.

As far as we know, plenty of methods attempt to explain vision-based RL agents from the perspective of feature attribution\cite{peeking,explainable_ai,concepts_algorithms}.
For example,  the embedding-based methods \cite{annasamy2019towards} analyze high-dimensional feature maps, and the attention-based methods \cite{manchin2019reinforcement} learn saliency maps in the agent's training process.
However, the former approach may be intuitively complex and the latter one does not apply to the post-hoc explanation.
Other methods that manage to discover the interpretable features in an easy-to-understand way without revising the DNN's architecture (such as the perturbation-based \cite{greydanus2018visualizing}, gradient-based \cite{zahavy2016graying}, and supervision-based \cite{shi2020self} methods)  generally build on the same action matching principle. 
Specifically, they usually attribute features of states by ensuring action consistency when explaining a pretrained RL agent.
However, we have some concerns about this commonly used principle since RL agents’ intrinsic goal is to achieve the maximum accumulated rewards instead of action cloning.

\begin{figure}[ht]
\centering
\includegraphics[width=\linewidth]{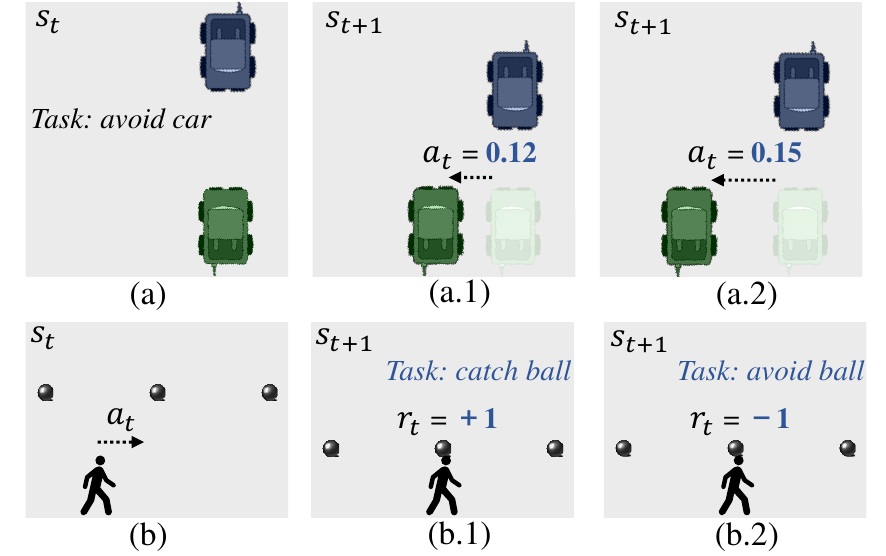}
\vspace{-20pt}
\caption{Examples of action matching's limitations. \textbf{(a)} Actions moving left by 0.12 and by 0.15 represent the same behavior ``avoiding collision'' and get the same reward. \textbf{(b)} The same action leads to different rewards when the tasks differ, but action matching methods would give the same~explanation.}
\label{3_1}
\vspace{-10pt}
\end{figure}

\IEEEpubidadjcol
Considering the situation when agents take slightly different actions but intrinsically aim at the same behavior to get a certain reward, the attentive features discovered by action matching can be redundant or even misleading. For example, as illustrated in Figure~\ref{3_1}(a), the agent in a driving task gets positive rewards when successfully avoiding oncoming cars. In the state $s_t$, the agent (the green car) needs to avoid collision with the oncoming blue car. However, the actions $a_t$ ``moving left by 0.12'' in Figure~\ref{3_1}(a.1) and ``moving left by 0.15'' in Figure~\ref{3_1}(a.2) would have obvious differences, while their essence is the same ``avoidance'' behavior. In this case, the behavior is truly represented by the reward instead of the action. Action matching may learn some redundant attention to keep the identical speeds or angles in the action, which can be seen as a kind of ``overfitting''.

Considering another situation when agents take the same action but get different rewards, the action matching methods may fail to discover the truly interpretable features. For example, as illustrated in Figure~\ref{3_1}(b), the agents take the same action ``moving right'' under the same state. Then the agent in the catching-ball task gets a positive reward, while the agent in the avoiding-ball task gets a negative reward. Intuitively, the agent in Figure~\ref{3_1}(b.1) may take the action because it intends to catch the middle ball, and the agent in Figure~\ref{3_1}(b.1) may move right because it attempts to avoid the left ball. However, the action matching methods would learn the same attentive features because of the same action, and the task-related explanation may not be truly discovered.

In this paper, we argue that action matching, the classical setting of RL explanation, may have limitations because it simply assumes that the agent's actions fully represent the underlying decision-making logic.
On the contrary, for deep RL, it is suggested that merely matching actions to attribute features of states is equal to explaining the DNN's forward process.
It barely focuses on the RL agents or the controlling tasks and only relates to the DNN's architectures and parameters.
Considering that the agent's training objective lies in the reward (or return) instead of the action, we assume that action matching cannot truly interpret the agent as it diverges from RL's reward-oriented motivation.
In fact, plenty of methods have realized the limitations of action-related explanation and attempt to understand RL agents by exploring causal effects on rewards~\cite{heuillet2021explainability}.
However, they are usually not as visually understandable as the feature attribution methods because of the structural causal equations.

Therefore, we attempt to take the causal effects into consideration when visualizing feature attribution. Unlike prior methods discussing how agents take actions, we aim to focus on the causal effect of actions and consider the agent's reward-driven behaviors.
Our proposed method leverages reward consistency to explore such causal effects, which means that the interpretation model needs to discover the most salient features in the state that affect the agent's obtained reward.
To solve the gradient disconnection from actions to rewards when the agent interacts with the environment, an RL framework is adopted to achieve reward consistency (such RL interpreting RL model is denoted as RL-in-RL).
Our main contributions are summarized as follows: 1) The causal effect on rewards during states' feature attribution is firstly explored; 2) A novel framework is proposed where the interpretation problem is modeled as a new RL task; 3) We analyze the limitations of the conventional action matching principle and introduce the concept of reward matching.
To substantiate these novel ideas and methodologies, we conduct comprehensive experiments. The results and detailed discussions underscore the significance and effectiveness of our proposed method, pointing out that the action matching principle in post-hoc explanations can lead to irrelevant and non-causal attention.

%% file: chapters/2_related_work.tex
In this section, we first review the mainstream methods of explaining vision-based RL agents.
Existing works that aim to discover how inputs influence agents' decisions can be partitioned into five main categories: embedding-based methods, attention-based methods, gradient-based methods, perturbation-based methods, and supervision-based methods.
Then, since this paper focuses on a post-hoc scenario in which an interpretation model learns the feature attribution of a pretrained and static policy, applications of these main methods for post-hoc explanations are summarily discussed.
Besides those works that explain RL agents mainly from the perspective of feature attribution, the causality-based interpretation methods that consider causal effects are also reviewed.

\textbf{Embedding-based methods.} Methods in this category generally focus on visualizing high dimensional data with a commonly used non-linear dimensionality reduction method, t-SNE \cite{van2008visualizing}. A simple idea is to directly draw a t-SNE map where perceptually similar states are demonstrated as points in the vicinity \cite{mnih2015human,zahavy2016graying,annasamy2019towards}. A recent work uses 1D t-SNE to design the DRLViz model \cite{jaunet2020drlviz}, which is able to re-order the memory (\textit{i.e.}, t-SNE projection of absolute values) and identify decisive sub-sets. In addition, the distances among points can also be related to the transition probabilities \cite{engel2001learning}.  However, t-SNE maps are intuitively complex for those without background knowledge of machine learning. 

\textbf{Attention-based methods.} Agents in this category learn to draw saliency maps without degrading their performance. Self-attention modules \cite{manchin2019reinforcement,nikulin2019free} which lay strong emphasis on relative areas of inputs are applied to augment the actor. Key-value attention structures \cite{choi2017multi,mott2019towards} are employed to learn interpretable policies by exploring how they look in the whole environment. Varieties of attention-based interpretation methods are newly proposed, such as TAFA \cite{zhang2021learning}, BR-NPA \cite{gomez2021improve} and Attentional Bottleneck \cite{kim2021towards}. However, these methods usually retrain the agent models and thus are inapplicable to structure-unchangeable or pretrained models. 

\textbf{Gradient-based methods.} The main idea underlying gradient-based methods is that the most salient features in the inputs contribute the largest gradient values. A typical approach is Jacobian saliency maps \cite{simonyan2013deep}, where spatial attributions in input states are computed as the Jacobian through back-propagation. In addition, there are several works modifying gradients to attain more meaningful and valuable saliency maps, such as LRP \cite{bach2015pixel}, DeepLIFT \cite{shrikumar2016not}, Smoothgrad \cite{smilkov2017smoothgrad}, Grad-CAM \cite{selvaraju2017grad}, and SEG-GRAD-CAM \cite{vinogradova2020towards}. Gradient-based methods uncover the dependencies between inputs and outputs simply and efficiently, but may not produce trustworthy interpretations as the manifold is changed \cite{greydanus2018visualizing}.

\textbf{Perturbation-based methods.} These methods aim to measure relative feature importance through perturbing the inputs, like applying occlusion masks and surrogating parts of the images. For example, RISE \cite{petsiuk2018rise} measures the feature importance by occluding inputs with random mask patterns, and Extremal Perturbation \cite{fong2019understanding} searches for a mask that exerts the most prominent impact on the outputs for a certain area. Recent works include FIDO \cite{chang2018explaining}, OPPSD \cite{kim2021one}, and PERT \cite{parvatharaju2021learning}. For inputs like images and videos, perturbation-based methods illustrate the effects of distinct parts in the form of saliency maps \cite{ivanovs2021perturbation,DBLP:journals/tsmc/YangWLD19}, but the illustration is usually coarse.

\textbf{Supervision-based methods.} This category of interpretation methods is first developed in supervised deep learning. These works aim to optimize for the salient regions using gradient descent techniques \cite{fong2017interpretable,dabkowski2017real,fong2019understanding}. On this basis, some research has been developed to interpret RL in such a supervision-based way, like SSINet \cite{shi2020self} and TSCI \cite{shi2021temporal}. The optimization goals of SSINet are action matching and attention sparsity, while TSCI explores temporal-spatial causal interpretations for the sequential decision-making process. In addition, a self-supervised method has been recently proposed to use visual attention as an inductive bias \cite{wu2021self}.

\textbf{Post-hoc Explanation Methods.} In vision-based RL tasks, the objective of post-hoc explanations is to identify the salient features within state inputs. Attention deemed valid underlines features that directly influence the behaviors of the agent.
Existing methods for post-hoc explanations predominantly fall into three categories: gradient-based, perturbation-based, and supervision-based approaches. Each carries its unique complexities and application conditions. The gradient-based methods involve computing the gradients of features that are notably prominent to the agent's current action output \cite{simonyan2013deep}. Perturbation-based methods assess the fluctuation in the action output following the perturbation or removal of certain input information \cite{van2008visualizing, fong2017interpretable}. Supervision-based methods employ an attention mask overlay on the input and instruct its action output to align with the original action at every step \cite{shi2020self}. It is worth noting that embedding-based methods, due to their high dimensionality, and attention-based methods, which function only during training, are less commonly used.
Fundamentally, these three methodologies are grounded in the assumption that the RL agent can be comprehensively interpreted via \textit{action matching}. This implies the identification of attention regions that significantly influence the agent's actions.

From the perspective of model understanding, matching actions is equal to matching the DNN's output in deep RL. Therefore interpretation methods built on action matching are factually explaining the DNN itself. It works well in supervised tasks when the DNN serves as a classifier, whose objective is to make its outputs imitate the true labels. However, in RL tasks, there is no label for action outputs. The DNN, as an RL agent, is trained in a reward-driven way. In this case, using action matching to interpret the agent does not coincide with its intrinsic goal since actions only indirectly illustrate the agent's reward-driven behaviors. It is, instead, the rewards that directly represent the agent's behaviors. In fact, plenty of research in RL causality has emphasized the role of rewards by exploring structural equations of states, actions, and rewards \cite{pearl2009causality,scholkopf2019causality,herlau2020causal,edmonds2018human,halpern2020causes}. Likewise, we believe that the exploration of reward-oriented effects can reasonably help post-hoc interpretation methods attend to more essential features. 

\textbf{Causality-based methods.} Causality is a crucial topic in the explanation of RL agents \cite{pearl2009causality,scholkopf2019causality,herlau2020causal}.
It remains unclear how agents can dynamically explore new environments and curtail the number of possibly feasible causal structures \cite{edmonds2018human}. Hence, it has become a popular topic whether RL agents can acquire causal knowledge. Structural causal models \cite{halpern2020causes}, assisted by structural equations, are usually adopted to represent the causal relationships. Furthermore, the Action Influence  \cite{madumal2020explainable} model, which includes actions in causal relationships, uses the state-action ensemble and structural equations to represent itself. 
To more effectively learn opportunity chains, the Distal Explanation  \cite{madumal2020distal} model utilizes a recurrent neural network for analysis and decision trees to promote the accuracy of the prediction. Another work named RAMi \cite{liu2021learning} is based on the Information Bottleneck \cite{tishby2000information} principle and aims at the Minimum Description Length.
Generally speaking, existing causality-based methods can reveal the inner causality but usually cannot be visualized in a user-friendly way when it comes to understanding vision-based RL agents.

%% file: chapters/3_preliminaries.tex
Reinforcement Learning (RL) refers to a general class of algorithms where the agent learns by interacting with the environment. Specifically, the agent takes an action $a_{t}$ in a state $s_{t}$ and receives a scalar reward $r_{t}$. Meanwhile, the environment $E$ changes to the next state $s_{t+1}$. The RL problem is generally modeled as a Markov decision process (MDP). An MDP can be described as a tuple $\mathcal{M} = (\mathcal{S}, \mathcal{A}, T, d_{0}, R, \gamma)$, where $\mathcal{S}$ is the state space, $\mathcal{A}$ is the action space, $T$ defines a transition function of the environment $T(s_{t+1}|s_{t}, a_{t})$, $d_{0}$ is the initial state distribution $d_{0}(s_{0})$, $R$ defines a reward function $R(s_{t}, a_{t})$, and $\gamma\in(0,1]$ is a discount factor~\cite{sutton1998introduction,levine2020offline}.

The final goal of an RL problem is to learn a policy, which defines a distribution over actions conditioned on states, $\pi(a_t|s_t)$.
The trajectory is a sequence of states and actions of length $H$, given by $\tau=\{s_0, a_0, \cdots, s_H, a_H\}$.
The trajectory distribution $p_{\pi}$ for a given MDP $\mathcal{M}$ and policy $\pi$ is:
\begin{equation}
p_{\pi}(\tau) = d_0(s_0)\prod_{t=0}^{H}\pi(a_t|s_t)T(s_{t+1}|s_t,a_t). 
\label{p_pi}
\end{equation}

The learned policy (\textit{i.e.}, the agent) aims at maximizing the return which means the expected sum of discounted future rewards.
The RL objective $J(\pi)$ over a horizon $H$ is:
\begin{equation}
J(\pi) = \mathbb{E}_{\tau \sim p_{\pi}(\tau)}\left [\sum_{t=0}^{H}\gamma ^{t}R(s_{t}, a_{t})\right ].
\label{J}
\end{equation}

In this paper, agents are trained with one of the mainstream methods, the proximal policy optimization (PPO) algorithm following~\cite{shi2020self}. As a policy gradient method in the actor-critic version, PPO uses trust region update to improve a general stochastic policy with gradient ascent \cite{schulman2017proximal}.

%% file: chapters/4_method.tex
In this section, we first present a reward-oriented interpretation method in Section~\ref{sec: Interpretation with Reward Consistency} to leverage reward consistency for interpretable feature discovery.
However, gradient disconnection from actions to rewards blocks its optimization. Hence, an RL-based framework that achieves reward consistency is further proposed in Section~\ref{sec: The RL-in-RL Model}. Last but not least, the extended version of our method for multi-step rewards (or return) consistency is presented in Section~\ref{sec: The RL-in-RL_k Model}.

\begin{figure*}[!t]
\centering
\includegraphics[width=0.93\linewidth]{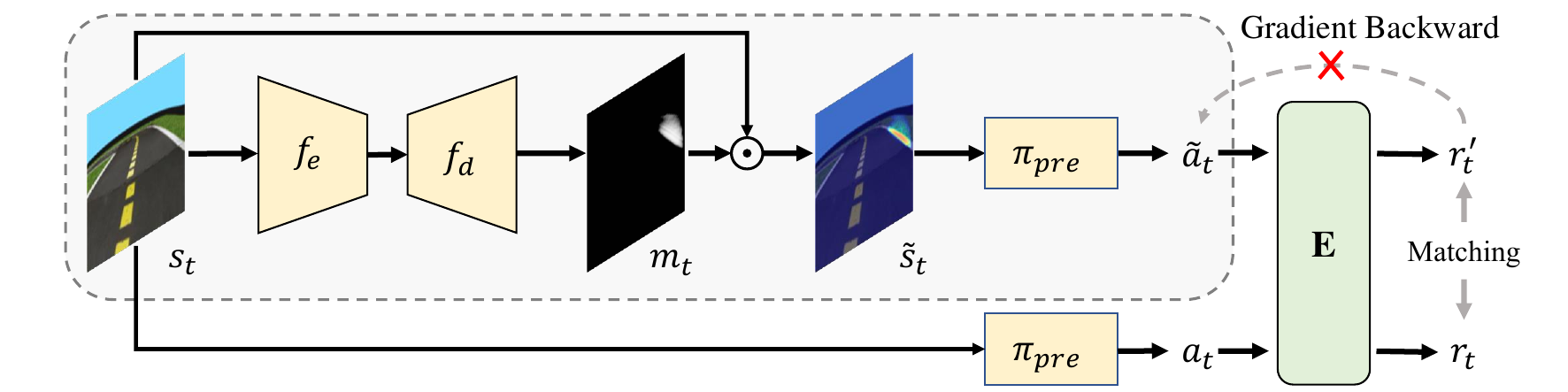}
\vskip -0.1in
\caption{\textbf{(1)} The gray box illustrates the architecture of our reward-oriented interpretation method. A DNN including an encoder $f_\text{e}$ and a decoder $f_{\text{d}}$ is adopted to learn the mask $m_t$ and attentive state $\tilde{s}_t$. \textbf{(2)} To conduct reward matching, the pretrained policy $\pi_{\text{pre}}$ takes the actions $a_t$ and $\tilde{a}_t$ respectively form the primitive state $s_t$ and attentive state $\tilde{s}_t$, and then the environment $E$ gives the corresponding rewards $r_t$ and $r_{t}'$. \textbf{(3)} However, the reward matching based on supervised learning is blocked by the disconnected gradient backward. Therefore, we model the reward matching problem as an RL task as in Figure~\ref{RL model}.} 
\label{arch}
\vspace{-10pt}
\end{figure*}

\subsection{Interpretation with Reward Consistency}
\label{sec: Interpretation with Reward Consistency}
Causality is the generic relationship between an effect and the cause that gives rise to it. For the interpretation problem, the state and policy can be taken as the ``cause'' while the reward is the ``effect''. 
Prior research explores causality mainly by the structural causal model (SCM) \cite{guo2020survey}. An SCM usually aims to construct a causal graph by structural equations. Causal effects in our interpretation problem can be described as $r = G(s|\pi,R)$. $G$ refers to a pre-defined causal structure that explicitly expresses the relationships among variables. However, such a causal structure is usually unknown, and designing it can be a great challenge especially under complex tasks or with ambiguous effects.   

As DNNs achieve superior performance in a variety of learning tasks with their powerful representation ability\cite{unet,semantic,depth_estimation}, we consider using a DNN to discover the causal effects. Note that the objective is $r = G(s|\pi,R)$, which can be seen as exploring what features in states affect the agent's obtained rewards, given the policy $\pi$ and reward function $R$. Hence, we design a mask-based architecture as an implicit causal structure to leverage reward consistency. 
As illustrated in Figure \ref{arch}, the encoder $f_{\text{e}}$ extracts high dimensional features of the input state $s_t$, and then the decoder $f_{\text{d}}$ learns a mask $m_t$. The mask displays the attentive feature importance for $s_t\in\mathbb{R}^{h\times w\times c}$, and $m_t\in[0,1]^{h\times w\times 1}$. In a frame, $h$ is the height, $w$ is the width, and $c$ is the channel. Next, multiplying the original state by the mask results in the attentive state $\tilde{s}_t$. We adopt the input state and the attentive state to get the action $a_t$ and the attentive action $\tilde{a}_t$ respectively by the pretrained policy. The environment gives rewards ($r_{t}$ and $r_{t}'$) according to different actions. The process can be formulated as follows:
\begin{subequations}\label{two_reward}
\setlength{\abovedisplayskip}{-1pt}
\begin{align}
r_{t} = R\big(s_t, \pi_{\text{pre}}(s_t)\big)\ \ ,\label{r}\\
r_{t}' = R\big(s_t,\pi_{\text{pre}}(\tilde{s}_t)\big)\ \ ,\label{r_bar}\\
\tilde{s}_t = s_t\cdot f_{\text{d}}\big(f_{\text{e}}(s_t)\big)\ \ .\label{s_bar}
\end{align}
\end{subequations}

We match $r_t$ and $r_{t}'$ to learn the attentive state which highlights features that directly give~rise to the current reward. The goal of this optimization problem is to ensure reward consistency by minimizing $|r_{t} - r_{t}'|$. 

An intuitive way to achieve this reward matching is the direct gradient backward as supervision-based action matching does.
Specifically, after the attentive state $\tilde{s}_t$ is obtained, action matching simply calculates a matching loss between the attentive action $\tilde{a}_t$ and the original action $a_t$. For example, the loss is $L = \left \| a_t - \tilde{a}_t \right \|_2$. The projection from states to actions through the pretrained policy is continuous and derivable. Thus, direct gradient backward works well. However, when it comes to rewards, such a matching way meets a setback, as shown in Figure \ref{arch}. The environment gives the reward when receiving the action. There exist internal transition functions in the environment which work in a black-box and end-to-end way. The reward function $R$ is unknown and non-differentiable, which makes the direct gradient backward from rewards to actions impossible. Therefore, the classical supervised learning method doesn't apply to this reward-related interpretation method. On the other hand, gradient disconnection between actions and rewards never hinders the optimization process in RL problems, which provides a feasible solution to our network's optimization problem. 

\subsection{The RL-in-RL Model}
\label{sec: The RL-in-RL Model}

Since the internal reward function $R$ is non-differentiable, reward matching cannot be conducted in the supervised learning way. Therefore, We model the optimization problem in Section \ref{sec: Interpretation with Reward Consistency} as a reinforcement learning task, which considers reward matching as part of the environment dynamic and thus avoids the gradient disconnection. Accordingly, a new policy is learned to interpret the pretrained policy via maximizing the newly designed rewards where the reward matching objective is implicitly incorporated. Such an RL interpreting RL model is referred to as the RL-in-RL model.

\begin{figure}[ht]
\centering
\includegraphics[width=0.95\linewidth]{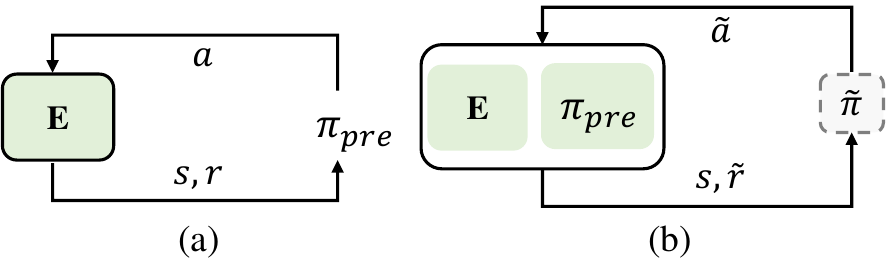}
\vspace{-10pt}
\caption{The interaction process. \textbf{(a)} illustrates how $\pi_{\text{pre}}$ interacts with the environment during pretraining. \textbf{(b)} illustrates the interpretation task where reward matching is modeled as an RL problem. The defined RL-in-RL policy $\tilde{\pi}$ corresponds to the gray box in Figure~\ref{arch}. The reward $\tilde{r}$ is given jointly by the environment $E$ and the pretrained policy~$\pi_{\text{pre}}$.}
\label{RL model}
\vspace{-10pt}
\end{figure}

Provided that the pretrained policy $\pi_{\text{pre}}$ is given, we denote its state space as $\mathcal{S}$, its action space as $\mathcal{A}$, and the environment as $E$.
Given the current state $s_{t}\in\mathcal{S}$, the action $a_{t}\in\mathcal{A}$ is deterministically decided by $a_{t} = \pi_{\text{pre}}(s_{t})$.
Note that both deterministic and stochastic policies are utilized to generate deterministic actions during the inference process, following the common practice for RL agents\cite{schulman2017proximal,sac,IQL,CQL}. 
The environment $E$ receives the action $a_{t}$ and gives the reward by its reward function $r_{t}=R(s_t, a_t)$. The goal of $\pi_{\text{pre}}$ is to get the maximum accumulated rewards from the environment $E$. Figure~\ref{RL model}(a) illustrates the RL interaction process of $\pi_{\text{pre}}$.

For the RL-in-RL model, our goal is to learn a new interpretation policy $\tilde{\pi}$. The state space $\mathcal{S}$ and action space $\mathcal{A}$ keep consistent with the pretrained policy $\pi_{\text{pre}}$.  
The RL-in-RL policy takes the action $\tilde{a}_t = \tilde{\pi}(s_t)$. For the current interpretation task, the goal changes to achieving the maximum consistency between rewards from the $a_{t} = \pi_{\text{pre}}(s_{t})$ and from the $\tilde{a}_t = \tilde{\pi}(s_t)$. That is to say, for the same state $s_t$ and a well-trained $\tilde{\pi}$, the environment $E$ gives the same reward when receiving the action decided by the interpretation policy $\tilde{\pi}$ as that when receiving the action decided by the pretrained policy $\pi_{\text{pre}}$. The new reward $\tilde{r}_t$ for the RL-in-RL model can be formulated as follows:
\begin{equation}
\begin{aligned}
\tilde{r}_t 
&= \tilde{R}\Big(s_t,\ \ \tilde{a}_t\Big) \\
&= -D\Big(R\big(s_t, a_t\big),\ \ R\big(s_t,\tilde{a}_t\big)\Big) \\
&= -D\Big(R\big(s_t,\pi_{\text{pre}}\left(s_t\right)\big),\ \ R\big(s_t,\tilde{\pi}\left(s_t\right)\big)\Big),
\label{new_r}
\end{aligned}
\end{equation}
where $D$, utilized as the Mean Squared Error (MSE) in our experiments, measures the distance between two variables. Figure~\ref{RL model}(b) illustrates the interaction process of RL-in-RL. Note that the current reward $\tilde{r}_t$ is jointly given by the environment $E$ as well as the fixed policy $\pi_{\text{pre}}$.
The gray box in Figure~\ref{RL model}b corresponds to $\tilde{\pi}$ in Figure~\ref{arch}, which~means:
\begin{equation}
\label{pi_tilde}
\tilde{\pi}(s_t) = \pi_{\text{pre}}\big(s_t\cdot f_{\text{d}}\big(f_{\text{e}}(s_t)\big)\big).
\end{equation}

When calculating $\tilde{r}_t$ as in Equation \eqref{new_r}, the input states of $\pi_{\text{pre}}$ and $\tilde{\pi}$ need to be exactly identical, otherwise matching the corresponding rewards becomes misleading. Meanwhile, minor differences between $a_t$ and $\tilde{a}_t$ can lead to huge pixel-wise differences in the next state $s_{t+1}$. It's hard to obtain the same actions from attentive states as from the original states, and exact action matching is not our goal either. Therefore, to keep the same input states of $\pi_{\text{pre}}$ and $\tilde{\pi}$ at every step $(s_t, \tilde{a}_t, \tilde{r}_t)$, we consider that one trajectory only contains one step in RL-in-RL. That is to say, the environment $E$ resets and gives initial states after every interaction. The objective of this RL task can be formulated as follows:

\begin{equation}
\begin{aligned}
J_{\text{RL}} = \mathbb{E}_{\pi}\left [\tilde{R}(s_{t}, \tilde{a}_{t})\right ].
\label{pi_in}
\end{aligned}
\end{equation}

Adopting RL algorithms to maximize the above objective ensures that the mask learns decisive features in states but may also lead to attention degradation. That is to say, all the features could share the same importance, \textit{i.e.}, the maximum attention value 1. In that circumstance, we can get the maximum $\tilde{r}$ but still have no idea what features matter. To solve such attention degradation, an auxiliary task is trained along with the RL interpretation task, in order to encourage the learned mask $m_t$ to be as sparse as possible. In other words, the number of features with high attentive importance needs to be possibly small. The objective of the auxiliary task is:

\begin{equation}
\label{J_aux}
J_{\text{aux}} = -\left \| m_t \right \|_1 = - \left \| f_{\text{d}}\big(f_{\text{e}}(s_t)\big) \right \|_1,
\end{equation}
where $\left \| \cdot \right \|_1$ denotes the $L_1$-norm. The final objective contains the RL term and auxiliary term, shown as follows:
\begin{equation}
\begin{aligned}
J = J_{\text{RL}} + \alpha J_{\text{aux}}
= \mathbb{E}_{\pi}\left [\tilde{R}(s_{t}, \tilde{a}_{t})\right ] - \alpha\left \| f_{\text{d}}\big(f_{\text{e}}(s_t)\big) \right \|_1,
\label{total_loss}
\end{aligned}
\end{equation}
where $\alpha$ is a positive scalar controlling the sparseness of the mask. The pseudo-code of training is summarized in Algorithm \ref{algorithm}. Note that the encoder in RL-in-RL shares the feature extractor parameters of the pretrained policy to speed up convergence and prevent overfitting \cite{shi2020self}.

\begin{algorithm}[H]
\caption{The RL-in-RL Model}\label{alg:algorithm}
\begin{algorithmic}
  \STATE The fixed pretrained policy $\pi_{\text{pre}}$ to be interpreted;
  \STATE The reward function $R(s_t, a_t)$ by environment $E$;
  \STATE Load $f_{\text{e}}$ and initialize $f_{\text{d}}$;
  \FOR{epoch = 0, 1, ... , until convergence}
    \FOR{$i$ = 1, 2, ... , $K$}
    \STATE Initialize state $s_0$;
    \STATE Get $\tilde{a}_0 = \pi_{\text{pre}}\Big(f_{\text{e}}\big(f_{\text{d}}(s_0)\big)\cdot s_{0}\Big)$;
    \STATE Get $\tilde{r}_0=-D\Big(R\big(s_0,\pi_{\text{pre}}\left(s_0\right)\big), R\big(s_0,\tilde{a}_0\big)\Big)$;
    \STATE Save $\tau^i = (s_0, \tilde{a}_0, \tilde{r}_0)$;
    \ENDFOR
    \STATE Update $f_{\text{d}}$ by PPO algorithm to maximize the objective $J$ in Equation \eqref{total_loss};
  \ENDFOR
\end{algorithmic}
\label{algorithm}
\end{algorithm}

\subsection{The RL-in-RL$^K$ Model}
\label{sec: The RL-in-RL_k Model}

As a first step towards using reward consistency to interpret RL agents by feature attribution, we mainly focus on the one-step reward matching for a fair comparison, since existing action matching methods are generally limited to one-step action matching.
Nevertheless, RL-in-RL is compatible with the one-step reward as well as multi-step rewards (or returns) consistency.
To adopt the multi-step version of RL-in-RL (denoted as RL-in-RL$^K$), the defined reward function in Equation~\eqref{new_r} is reformulated as~follows:

\begin{equation}
\label{multistep}
\begin{aligned}
\tilde{r}_{t}^{K} 
= -D\Big(\sum_{i=0}^{K-1}\gamma^{i}R\big(s_{t+i}, a_{t+i}\big),\ \ \ \ \ \ \ \ \ \ \ \ \ \ \ \ \ \  \\
R\big(s_t,\tilde{a}_t\big) + \sum_{j=1}^{K-1}\gamma^{j}R\big(s_{t+j}^{'}, a_{t+j}^{'}\big)\Big),
\end{aligned}
\end{equation}
where $s_{t+i+1}=T(s_{t+i},a_{t+i})$ and $a_{t+i}=\pi_{\text{pre}}(s_{t+i})$ are in the pretrained behavior trajectory, $s_{t+1}^{'} = T(s_{t},\tilde{a}_{t})$, $s_{t+j+1}^{'} = T(s_{t+j}^{'},a_{t+j}^{'})$, and $a_{t+j}^{'}=\pi_{\text{pre}}(s_{t+j}^{'})$ are in the interpretation behavior trajectory, $\gamma$ is a discount factor, and $K$ is the observation length.

When $K=1$, it equals the interpretation problem with one-step reward consistency, \textit{i.e.}, RL-in-RL.
When $K>1$, the RL-in-RL$^{K}$ focuses on observing the long-term effects of the behavior, and keeps the multi-step rewards or even return (\textit{$K=H$}) consistency.
By lengthening the step of reward consistency (\textit{i.e.}, adopting a larger observation length $K$), the attentive features would illustrate more information that helps explain long-term behaviors.

%% file: chapters/5_1experiments.tex
In this section, we first describe experimental settings and adopted environments in Section~\ref{sec: Setup}. Then, results are shown to validate RL-in-RL for its interpretability in Section~\ref{sec: Evaluations}. 

\subsection{Setup}
\label{sec: Setup}
Our experiments are conducted on the Atari 2600 games and the Duckietown environment. The Euclidean Distance is adopted as the $D$ function in Equation \eqref{new_r} and the UNet \cite{DBLP:conf/miccai/RonnebergerFB15} serves as the encoder-decoder architecture in Figure \ref{arch}. Other experimental details and ablation studies on our hyperparameters are in the Appendices.

\textbf{Atari 2600} \cite{bellemare2013arcade} is a widely used benchmark in the field of RL interpretation. It consists of various RL tasks like playing ping-pong and hiding from monsters. Actions in these tasks are all discrete such as ``left'', ``right'', ``up'', and ``down''. The states are color images and we adopt the 84$\times$84$\times$4 stacked grayscale images as inputs following \cite{mnih2015human}. The reward during training is normalized to -1, 0, and 1.

\textbf{Duckietown} \cite{gym_duckietown} is an autonomous driving simulator environment. 
The states are 120$\times$160$\times$3~color images from a single camera. The actions contain two continuous numbers between -1 and 1, corresponding to the forward velocity and steering angle. A positive velocity makes the agent go forward, and a positive steering angle makes it turn left. The reward function consists of the speed reward, the lane reward, and the obstacle penalty. The speed reward term encourages a large driving speed. The lane reward requires the agent to drive on the right side of the lane. The obstacle penalty asks the agent to avoid obstacles and invalid driving areas (for example, the grass outside the lane). 

\begin{figure*}[!t]
\centering
\includegraphics[width=0.93\linewidth]{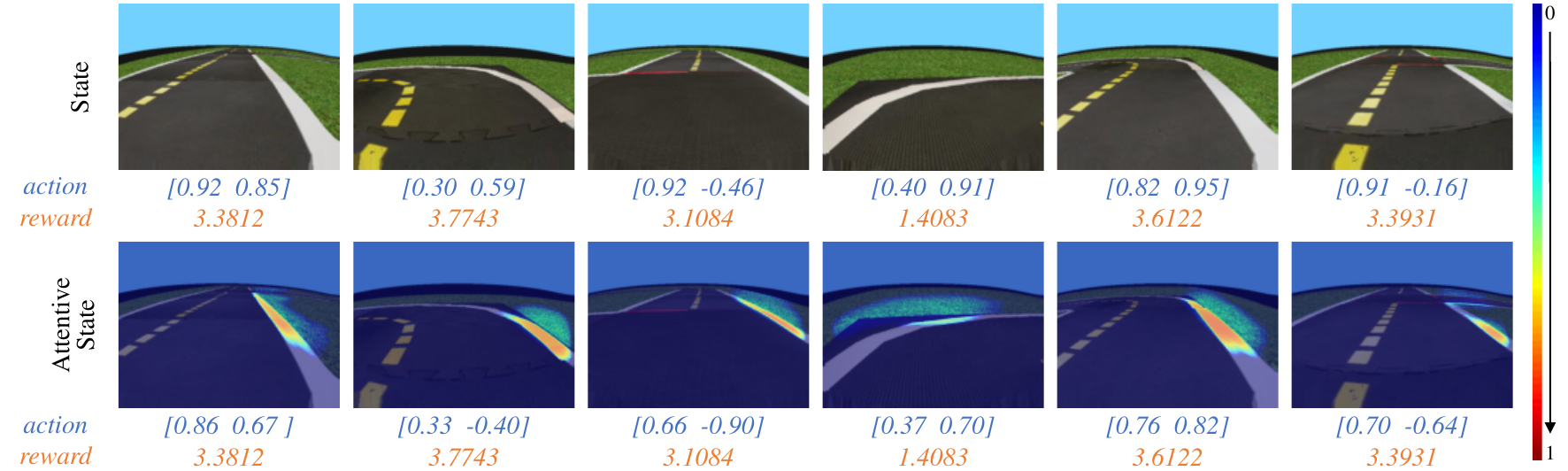}
\vspace{-5pt}
\caption{The performance of our proposed RL-in-RL model in the Duckietown environment. The attentive state is an overlaid combination of the state and the attentive heatmap of feature attribution. The feature importance ranges from 0 to 1 as the heatmap color changes from blue to red.}
\label{5_1}
\vspace{-10pt}
\end{figure*}

\begin{figure}[t]
\centering
\includegraphics[width=0.93\linewidth]{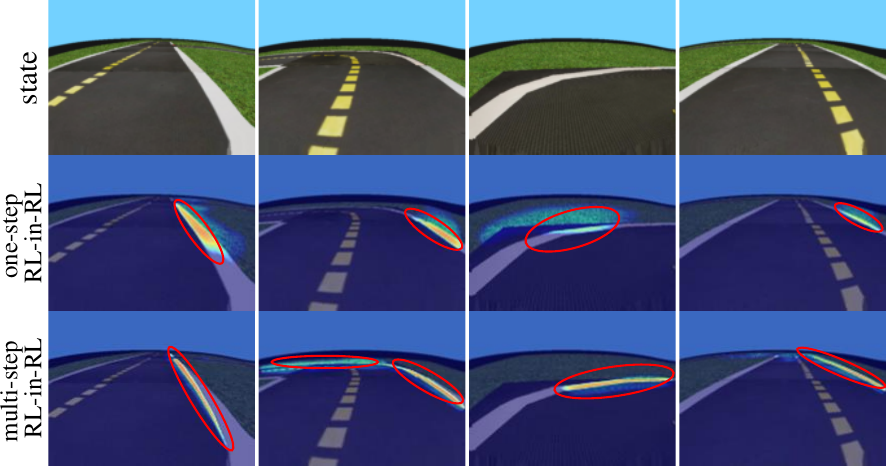}
\vspace{-5pt}
\caption{The attentive features of RL-in-RL$^K$ with observation length $K=10$.}
\label{multistep}
\vspace{-10pt}
\end{figure}          

\begin{figure*}[!t]
\centering
\includegraphics[width=0.95\linewidth]{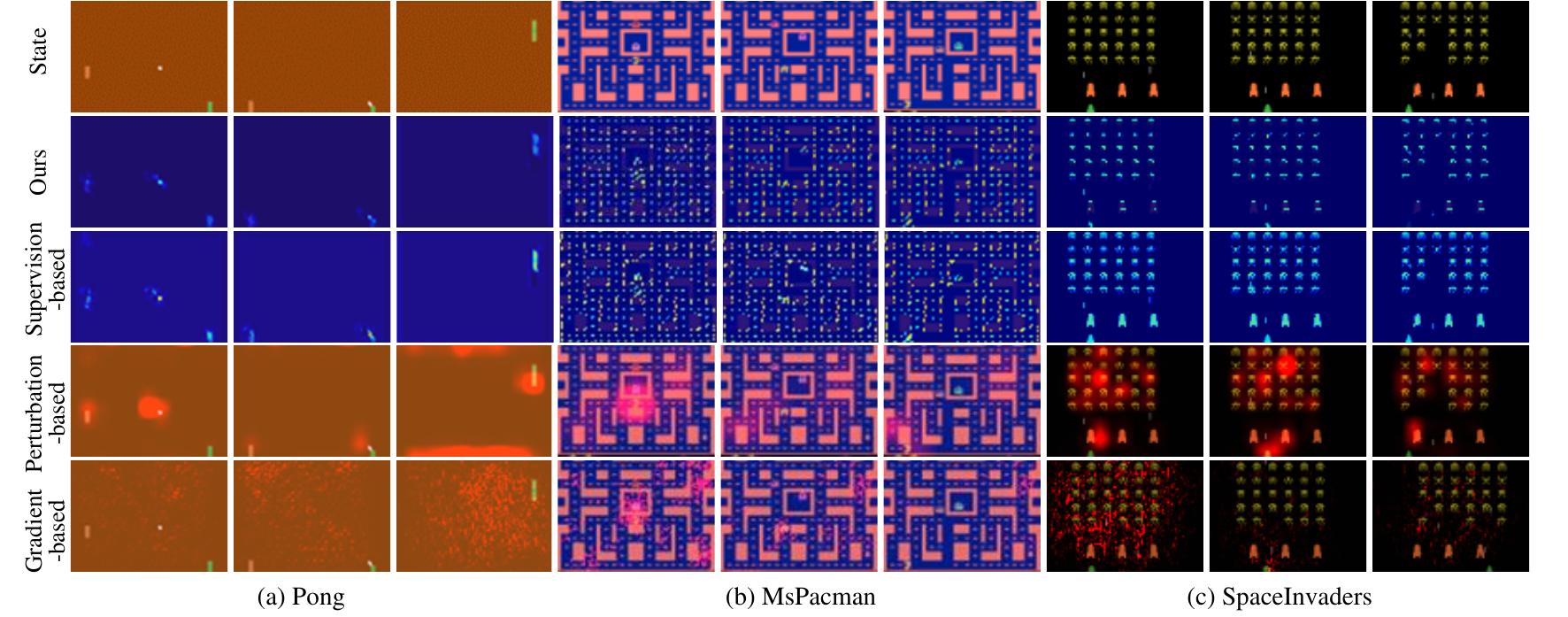}
\vspace{-10pt}
\caption{Comparisons among different interpretation methods on Atari 2600. Our RL-in-RL model and the supervision-based method are visualized in the overlaid heatmap as in Figure \ref{5_1}. The perturbation-based and gradient-based methods highlight the attention areas on the saliency-overlaid state.}
\label{5_2}
\vspace{-5pt}
\end{figure*}

\subsection{Evaluations}
\label{sec: Evaluations}
We investigate whether the proposed method provides valid explanations and compare RL-in-RL with other popular approaches.
Specifically, experiments in this section aim at answering two questions. \textbf{First}, does RL-in-RL achieve reward consistency as expected? \textbf{Second}, can RL-in-RL result in high-quality feature attribution for RL agents compared to action matching methods?
We emphasize that this work primarily focuses on validating reward consistency for interpretable feature discovery (Section~\ref{sec: Validity of Our Method}) and studying differences between reward matching and action matching (Section~\ref{sec: Understanding reward and action matching}). Our aim is not to ``beat'' all action matching methods or present a perfect method for RL~interpretation.

\textbf{Validity of RL-in-RL.} We compare the actions, rewards, and RL-in-RL's learned attention in the challenging self-driving environment.
As shown in Figure \ref{5_1}, the attentive state illustrates an overlaid state with its attentive heatmap of feature attribution. Quantitative results show that RL-in-RL manages to keep the same reward as that of the policy to be interpreted, and thus achieves our motivation to discover the reward-related features during interpretation. Note that actions are factually varied for the same rewards. It empirically proves our assumption that the intrinsic goal of a policy, the pursuit of rewards, cannot be fully represented by action matching.  Meanwhile, visualized results show that the pretrained policy mainly focuses on the right white line but attends to the near left white line when other lines are unobservable, which conforms to human attention under the right-hand traffic rule.

In addition to one-step reward consistency, we validate the proposed method with multi-step rewards consistency in Figure~\ref{multistep}. The RL-in-RL$^{K}$ generally maintains the same attention pattern (\textit{i.e.}, the left and right white lines) with one-step RL-in-RL, but the sights are set further ahead.
It is suggested that RL-in-RL$^K$ can not only leverage the reward information but also provide an adjustable observation length for long-term behavior interpretation.

\textbf{Comparative Evaluation.} Several popular RL explanation methods based on gradient \cite{zahavy2016graying}, perturbation \cite{greydanus2018visualizing}, and supervision \cite{shi2020self} are compared with our proposed method. 
The first metric is the saliency map, one of the most well-known visual methods where the intensity of the pixel color at a particular location corresponds to the importance of the input value~\cite{Agneza2022Explainability,The_Need_for_Interpretable_Features,he2021explainable}.
As illustrated in Figure~\ref{5_2}, the RL-in-RL and supervision-based methods learn more precise attention than that of the perturbation-based and gradient-based methods.
The former two methods also display the relative feature importance in the heatmap visualization while the latter two are incapable of telling. 
The second metric is the average return of the policy that only takes masked states as inputs, which quantitatively evaluates the quality of masks generated by various explanation methods.
In Table~\ref{tab:performance of baselines}, the RL-in-RL and supervision-based methods achieve comparable performance while the other two methods lead to significant performance degradation, when the policy can only access the attentive pixels learned by corresponding explanation methods.

\begin{table}[ht]
\caption{Average returns over 5 random seeds}
\label{tab:performance of baselines}
\centering
\begin{tabular}{|p{14.5mm}<{\centering}|c|c|c|c|}
\hline
Task & Ours & \makecell[c]{Supervision\\method\cite{shi2020self}} & \makecell[c]{Perturbation\\method~\cite{greydanus2018visualizing}} & \makecell[c]{Gradient\\method\cite{zahavy2016graying}} \\
\hline
Enduro & 2802.43 & 2755.42 & 1741.10 & 819.92 \\
Seaquest & 2560.77 & 2356.67 & 1830.00 & 836.00 \\
SpaceInvaders & 738.46 & 740.45 & 562.92 & 297.73 \\
\hline
\end{tabular}
\end{table}

\begin{figure}[t]
\centering
\includegraphics[width=0.95\linewidth]{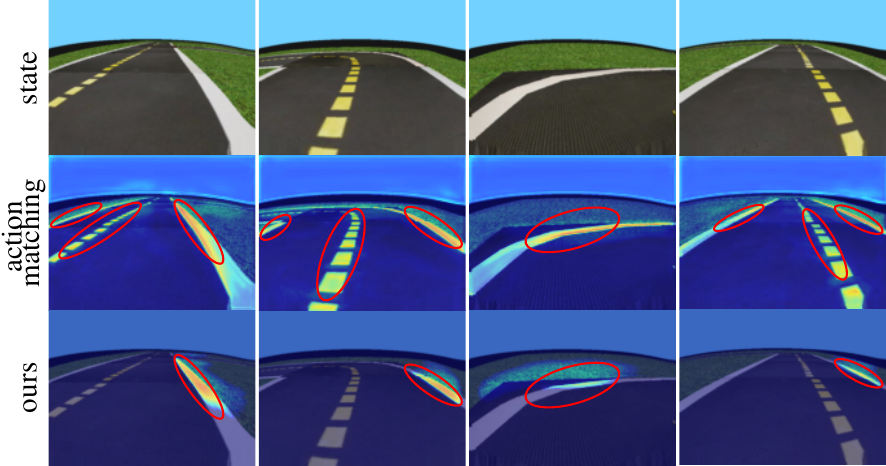}
\caption{Comparisons between the best-performing action matching method and our reward-oriented RL-in-RL on the Duckietown environment. 
}
\label{difference}
\vspace{-5pt}
\end{figure}

The supervision-based method with action matching has competitive precision with the RL-in-RL model but some attentive details are different.
However, such differences are subtle and analyses may be subjective. 
Therefore, we further compare the supervision-based action matching method with our reward-oriented method in the Duckietown environment.
As shown in Figure \ref{difference}, the main attention regions by action matching are the left white line, the middle yellow line, and the right white line (denoted as $\mathrm{F_{w}^{left}}$, $\mathrm{F_{y}}$, and $\mathrm{F_{w}^{right}}$ respectively). The results conform to our intuition since such three parts exactly constitute the whole lane. On the other hand, the RL-in-RL model merely highlights the right white line $\mathrm{F_{w}^{right}}$ except that if only the left white line is in sight (as shown in the third column of Figure \ref{difference}). Since the widths of the lane and the car are constants in Duckietown, the location of the right line can uniquely determine the locations of other lines. Hence, the attentive pattern discovered by RL-in-RL coincides with the task's reward function, where the agent is rewarded for driving on the right side of the lane and penalized for driving in the grass outside. 
To further understand their attention differences, a series of analytical experiments are designed in Section~\ref{sec: Understanding reward and action matching}.

%% file: chapters/5_2experiments.tex
In this section, we aim to understand reward matching and action matching while interpreting RL agents by exploring their attention differences in Figure~\ref{difference}.
The fine-grained action matching method based on the supervised learning \cite{shi2020self} is compared with RL-in-RL in the Duckietown environment. 
Since the right white line $\mathrm{F_{w}^{right}}$ always has high importance in the above two explaining patterns in Figure~\ref{difference}, our main concern is whether the left white line $\mathrm{F_{w}^{left}}$ and the middle yellow line $\mathrm{F_{y}}$ are redundant attention. Hence, to explore truly crucial features of the pretrained policy, experiments are designed to answer the following three questions.

\begin{figure*}[ht]
\centering
\subfloat[]{\includegraphics[width=0.5657\linewidth]{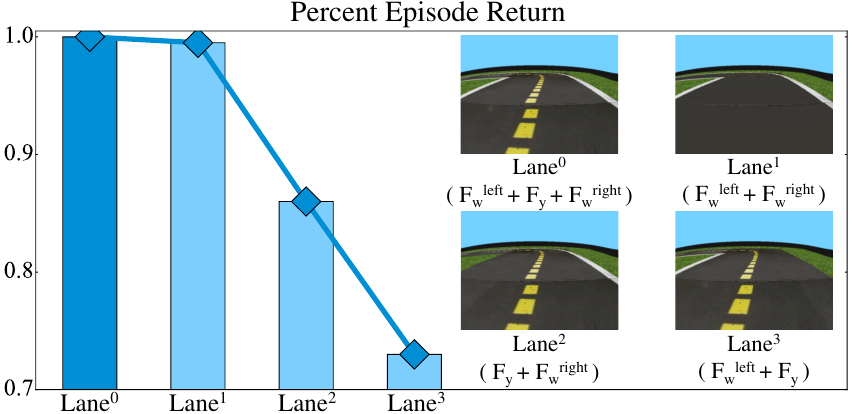}%
\label{lane_patterns}}
\hfil
\subfloat[]{\includegraphics[width=0.4243\linewidth]{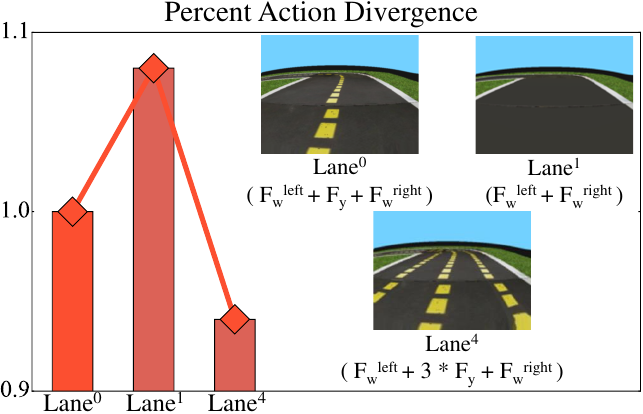}%
\label{kl_divergence}}
\caption{Quantitative analyses averaged across 100 random seeds. \textbf{(a)} The percent episode return of the pretrained policy under different lane patterns. This measures how three lines (\textit{i.e.}, $\mathrm{F_{w}^{left}}$, $\mathrm{F_{y}}$, and $\mathrm{F_{w}^{right}}$) affect the policy's performance respectively. \textbf{(b)} The percent action divergence of the interpretation model under different lane patterns, compared to the pretrained policy's actions. This measures how three lines affect the interpretation model's action consistency with the pretrained policy respectively.}
\label{lane_and_kl}
\end{figure*}

\begin{figure}[t]
\centering
\includegraphics[width=0.95\linewidth]{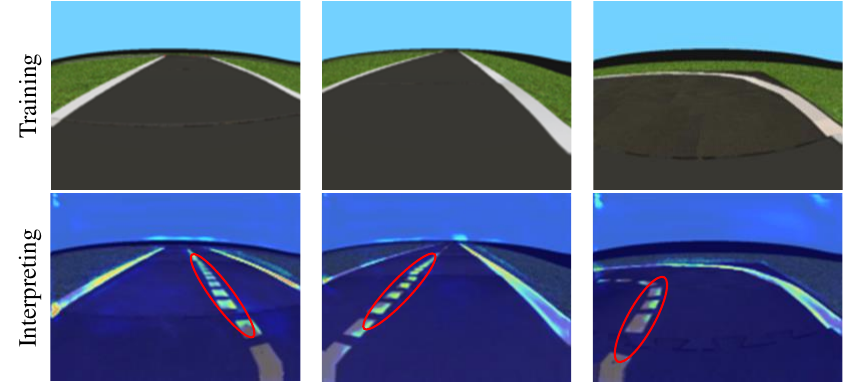}
\caption{The interpretation results from the action matching method, when the policy is pretrained without the middle yellow line.}
\label{b.1}
\vspace{-5pt}
\end{figure}

\begin{figure}[t]
\centering
\includegraphics[width=0.98\linewidth]{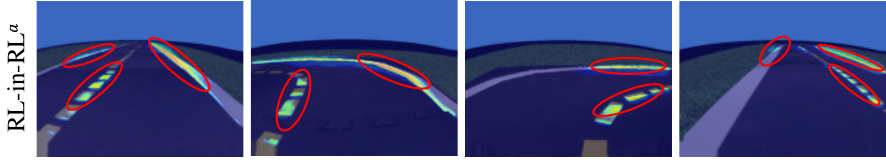}
\caption{The attention pattern of RL-in-RL$^a$.}
\label{RL-in-RL'}
\vspace{-5pt}
\end{figure}

\textbf{1) Is the extra attention discovered by the action matching method redundant?}

The first question is whether the extra attention $\mathrm{F_{w}^{left}}$ and $\mathrm{F_{y}}$ are redundant or not. In other words, do they truly affect the performance of the pretrained policy? To figure out this question, the pretrained policy is evaluated under different lane patterns as shown in Figure~\ref{lane_patterns}. The lane$^0$ is the primitive pattern that the policy is pretrained with, which has two white lines and one yellow line. 

Experimental results are illustrated in Figure~\ref{lane_patterns}. Firstly, $\mathrm{F_{y}}$ is removed in the lane$^1$, and the pretrained policy maintains the comparative performance. It validates that the yellow line is not the essential feature of the pretrained policy and thus the attention is redundant. Secondly, $\mathrm{F_{w}^{left}}$ is removed in the lane$^2$, and the pretrained policy's performance slightly decreases. In fact, the RL-in-RL model does highlight this region in cases where no other lines are in sight. Therefore, we suggest that the left white line occasionally affects the agent's decision but is redundant most of the time. 
Thirdly, $\mathrm{F_{w}^{right}}$ is removed in the lane$^3$, and a sharp decrease in performance is observed. It proves that the right white line is crucial for the agent and the RL-in-RL model discovers the truly essential feature. 

To further verify that the yellow line is the redundant attention, the policy $\pi_{\text{pre}}$ is re-trained in the lane$^1$ as shown in Figure \ref{lane_patterns}. Since the agent is trained without any involvement of the yellow line, its decision-making process obviously has nothing to do with this feature. We adopt the supervision-based action matching method to figure out its attention features and the results are illustrated in Figure \ref{b.1}. The interpretation method based on action matching still focuses on the middle yellow line, which further proves that action matching leads to irrelevant and redundant feature attention.  

\textbf{2) Does the action matching principle lead to redundant attention?}

After $\mathrm{F_{y}}$ is proved to be redundant attention of the action matching method, the following question is what the cause is. It remains unclear whether it comes from the action matching principle or the supervised learning approach. To decompose these two factors, a variant of RL-in-RL (denoted as RL-in-RL$^{a}$) is designed to conduct action matching in the RL framework.
RL-in-RL$^{a}$ changes the reward function in Equation~\eqref{new_r} to:
\begin{equation}
\begin{aligned}
\tilde{r}_{t}^{a} 
= -D(a_t, \tilde{a_t})
= -D\big(\pi_{\text{pre}}\left(s_t\right), \tilde{\pi}\left(s_t\right)\big).
\label{RL-in-RL_v2}
\end{aligned}
\end{equation}

The goal of RL-in-RL$^{a}$ is the same as that of action matching, while its optimization method is reinforcement learning instead of supervised learning. Figure \ref{RL-in-RL'} shows that attentive regions in RL-in-RL$^{a}$ are similar to that in the supervision-based action matching method (as shown in Figure~\ref{difference}), where the redundant attention $\mathrm{F_{y}}$ still has high importance. In this case, we can conclude that the redundant attention results from the action matching principle instead of the optimization~method.

\textbf{3) Why does the action matching principle lead to redundant attention?}

As far as we know, the action matching principle can lead to fake attention like $\mathrm{F_{y}}$. Here comes the next question: why does it cause fake attention? In other words, what is the contribution of $\mathrm{F_{y}}$ during the process of matching actions? Considering a deterministic environment where the same action $a_t$ under the same state $s_t$ results in the same next state $s_{t+1}$, matching actions factually equals matching the next states. In this case, a minor variance in the agent's position can induce significant pixel differences in the next state. Hence, an assumption of the yellow line's role is to help the agent locate its current position and thus take the same action. To verify our assumption, another two yellow lines are added to the lane, as shown in lane$^4$ of Figure~\ref{kl_divergence}. Actions of one trajectory are collected separately for lane$^0$, lane$^1$, and lane$^4$. The action matching loss between two trajectories collected respectively under the original states and the attentive states for the same lane pattern is calculated as follows:
\begin{equation}
\label{kl_equation}
L_{\text{KL}} = D_{\text{KL}}(A_{s}, A_{\tilde{s}}),
\end{equation}
where $D_{\text{KL}}$ refers to the KL divergence and $A$ is the action trajectory within 500 interactions with the environment. As shown in Figure~\ref{kl_divergence}, compared to actions under the primitive pattern (\textit{i.e.}, lane$^0$), actions under the lane$^1$ have larger divergence while those under the lane$^4$ are less divergent. That is to say, extra attention on the yellow lane indeed raises the action matching degree although it's irrelevant to the pursuit of high rewards for the agent. Therefore, the redundant attention caused by action matching can be seen as a kind of ``overfitting'', which is dedicated to identical actions but neglects the reward-related goal of the policy.  

\textbf{Discussion.} What we know so far is that using action matching to explain RL agents would lead to redundant feature attribution. It results from the fact that actions only indirectly represent the agent's reward-related goal. On the other hand, we have to admit that interpretable features discovered by action matching methods indeed help users predict the agent's actions, which is a natural benefit of matching actions. Meanwhile, the interpretable features discovered with reward consistency are more like the ``safety boundary'', which are the key elements that guarantee the agent's expected performance. Hopefully, reward and action matching principles can complement each other and provide a more comprehensive understanding of deep RL in real-world applications, for both users and researchers.

%% file: chapters/7_conclusion.tex
In this paper, we discussed the limitations of the commonly used assumption, the action matching principle, in RL interpretation methods. 
It is suggested that action matching cannot truly interpret the agent since it differs from the reward-oriented goal of RL. 
Hence, the proposed method firstly leverages reward consistency during feature attribution and models the interpretation problem as a new RL problem, denoted as RL-in-RL.
Moreover, it provides an adjustable observation length for one-step reward or multi-step reward (or return) consistency, depending on the requirements of behavior analyses.
Extensive experiments validate the proposed model and support our concerns that action matching would lead to redundant and non-causal attention during interpretation since it is dedicated to exactly identical actions and thus results in a sort of ``overfitting''.
Nevertheless, although RL-in-RL shows superior interpretability and dispenses with redundant attention, further exploration of interpreting RL tasks with explicit causality is left for future work.

%% file: chapters/8_appendixA.tex
\subsection{Network Structure}
\label{sec: Network Structure}

\begin{figure}[ht]
\centering
\includegraphics[width=0.95\linewidth]{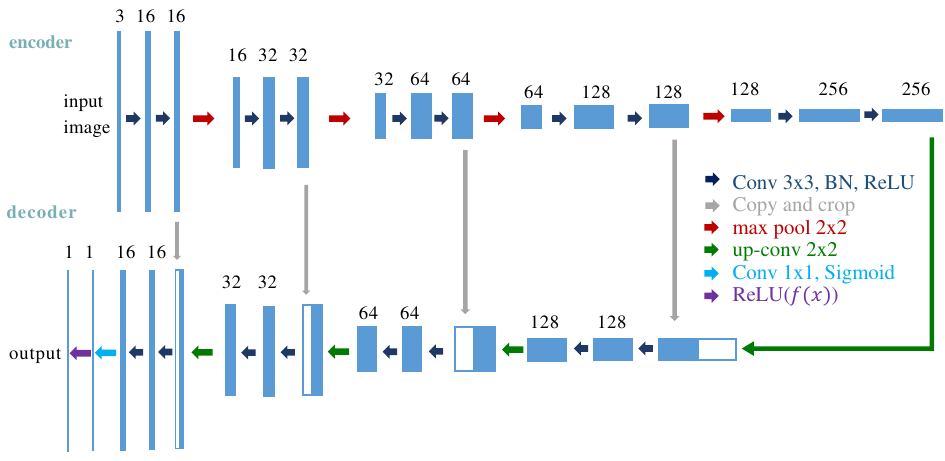}
\vspace{-8pt}
\caption{The architecture of the encoder-decoder network in RL-in-RL.}
\label{unet}
\end{figure}

In our experiments, the PPO algorithm is adopted for training. In PPO, the critic network is used to predict the state-value function and shares a common feature extractor with the actor. The actor and critic networks are each connected with a linear layer after the feature extractor. In the proposed method, the critic's linear layer of the pretrained policy is retrained because the state value changes in the interpretation RL problem. 
The RL-in-RL model, illustrated in Figure \ref{arch}, uses an encoder-decoder architecture with DNNs for a universal semantic segmentation task. It takes state images as inputs and produces learned saliency masks, a dense prediction task common in vision applications such as semantic segmentation \cite{semantic,unet} and scene depth estimation \cite{depth_estimation}. To mitigate task-irrelevant attention that may arise from complex architectures, we've adopted the simpler U-Net architecture with minor modifications, following its proven efficacy in RL explanations \cite{shi2020self,shi2021temporal}.
As shown in Figure \ref{unet}, firstly, the Sigmoid layer is added to project the mask from $\mathbb{R}^{H\times W\times C}$ to $[0,1]^{H\times W\times C}$, which can better reflect the relative feature importance and ensure the scale of attentive states. Secondly, the $ReLU(f(x))$ layer is added to speed up convergence. It can be formulated as $ReLU(\frac{x-\beta}{1-\beta})$, where $\beta\in[0,1)$ is a hyperparameter adjusting its effects on the speed of model converging. 

\subsection{Implementation}
\label{sec: Implementation}

We use the following software versions: Python 3.7, Pytorch 1.10.0~\cite{paszke2019pytorch}, Gym 0.15.4~\cite{brockman2016openai}, and Duckietown 6.2.24~\cite{gym_duckietown}.
Our implementations of gradient-based \cite{zahavy2016graying}, perturbation-based \cite{greydanus2018visualizing}, and supervision-based \cite{shi2020self} methods follow their respective author-provided implementations from GitHub or published papers.
The Optimal hyperparameters are adopted for each method.

%% file: chapters/8_appendixB.tex
\subsection{Validity of RL-in-RL}
\label{sec: apeendixB Validity of RL-in-RL}

\begin{figure*}[!t]
\centering
\includegraphics[width=\linewidth]{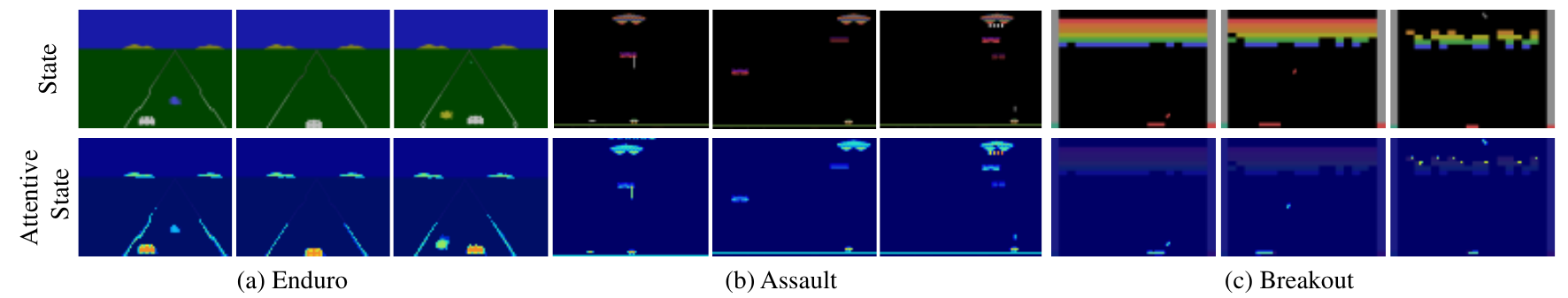}
\vspace{-20pt}
\caption{More results of the RL-in-RL model on Atari2600. The attentive states are visualized as in Figure \ref{5_1}.}
\label{b.2}
\end{figure*}

The proposed RL-in-RL model is verified on more tasks of the Atari2600, as shown in Figure \ref{b.2}. In the Enduro task for avoiding obstacles while driving, the attention is mainly on the driving boundary, obstacles, and the agent's own position. It is clear that the distant boundary is less attended to since it has a smaller impact on the current behavior. Similarly, the most distant obstacles are not noteworthy as shown in the third scenario. In the Assault task for offense and defense, attentive areas are generally the player, spaceships, and bullets. In the Breakout task for hitting colored squares with a ball, the agent mainly focuses on the small ball and the board, since whether the board catches the ball or not directly determines the reward. Another thing worth noting is the top of the rainbow block in the third scenario. It is suggested that for the pretrained policy, the actor prefers to send the ball to the top of the rainbow block to get a higher reward.

\subsection{Ablation Studies on Hyperparameters}
\label{sec: Ablation Studies on Hyperparameters}

\begin{figure}[ht]
\centering
\includegraphics[width=\linewidth]{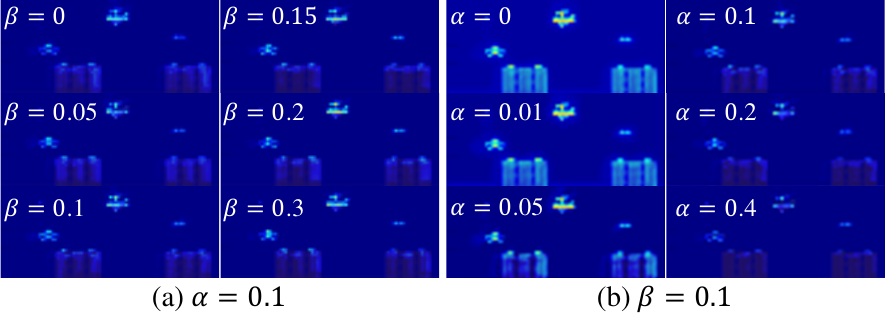}
\vspace{-15pt}
\caption{Two ablation studies on the hyperparameters. $\alpha$ is the weight of the auxiliary task loss as in Equation \eqref{total_loss}, while $\beta$ is to speed up the training process as described in Appendix \ref{appendixA}.}
\label{ablation}
\vspace{-5pt}
\end{figure}

The effects of hyperparameters $\alpha$ and $\beta$ are explored on the Atari AirRaid task. $\alpha$ is the weight of the auxiliary task loss as illustrated in Equation \eqref{total_loss}, and $\beta$ is to speed up the training process as described in Appendix \ref{sec: Network Structure}. Results are demonstrated in Figure \ref{ablation}. In the first experiment, $\alpha$ is fixed to 0.1, and $\beta$ is adjusted between 0 and 0.3. Visualized results are mostly the same since $\beta$ only affects the converging speed. In the second experiment, $\beta$ is fixed to 0.1 and $\alpha$ is adjusted between 0 and 0.4. 
When $\alpha$ is set to 0, the attention almost highlights all of the observations. 
As $\alpha$ increases, the mask becomes more sparse and emphasizes more noteworthy parts.
Specifically, the background's and the buildings' attention importance gradually becomes lower because of their relatively small impact on the agent’s decision. The player's and aircraft's attention importance has become smaller but remains the most salient in the whole observation, as they directly affect the agent's obtained rewards.

\begin{figure}[ht]
\centering
\includegraphics[width=0.95\linewidth]{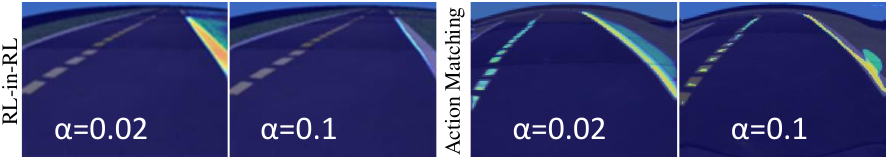}
\caption{Comparative results of the RL-in-RL method and the action matching method's ablation studies on the hyperparameter $\alpha$.}
\label{ablation_compar}
\vspace{-10pt}
\end{figure}

Furthermore, we compare the effect of $\alpha$ to the action matching method~\cite{shi2020self}.  This hyperparameter controls attention sparseness. As shown in Figure~\ref{ablation_compar}, the attentive patterns and their differences maintain as $\alpha$ varies.
It coincides with our conclusion in Section~\ref{sec: Understanding reward and action matching} that the redundant feature discovery is caused by the action matching principle instead of the optimization method or the hyperparameter. 

\subsection{Tasks with Sparse Rewards}
\label{sec: Tasks with Sparse Rewards}

This paper hasn't particularly discussed the sparse reward problem as most of the explainable RL research~\cite{heuillet2021explainability,glanois2021survey,chatzimparmpas2020survey,alharin2020reinforcement}. However, our experiments actually validated the ability of RL-in-RL to deal with such tasks. Specifically, in the Atari Pong, Breakout, and SpaceInvaders environments (cf. Figure~\ref{5_2} and \ref{b.2}), agents only get non-zero (1 or -1) rewards when the round is over. As shown in Table~\ref{tab:sparse reward}, during the game, only 1.36\%, 4.37\%, and 3.59\% of rewards are non-zero respectively (evaluated on 100 random seeds). Furthermore, the RL-in-RL$^K$ presented in Section~\ref{sec: The RL-in-RL_k Model} is compatible with the sparse reward problem and can fully leverage the sparse reward information.

\begin{table}[ht]
\vspace{-10pt}
\caption{Tasks with Sparse Rewards}
\label{tab:sparse reward}
\centering
\begin{tabular}{|c|c|c|c|}
\hline
Task & Pong & Breakout & SpaceInvaders \\
\hline
Horizon Steps & 1633.93 & 1356.50 & 921.09\\
Non-zero Reward Steps & 22.27 & 59.29 & 33.03 \\
Non-zero Reward Percentage & 1.36\% & 4.37\% & 3.59\% \\
\hline
\end{tabular}
\end{table}

\begin{figure}[ht]
\centering
\includegraphics[width=0.95\linewidth]{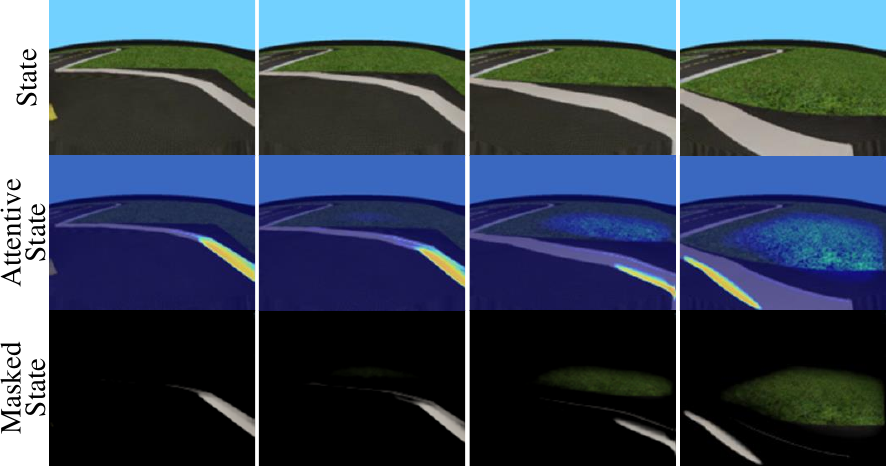}
\vspace{-4pt}
\caption{Visulization of attention shift in the zigzag turn. Three rows correspond to original states, attentive states, and masked states, respectively.}
\label{fig: fcase}
\end{figure}

\subsection{Analysis of Failure Case}
RL-in-RL provides an explanation for why RL agents may not perform robustly in novel situations different from their training scene from the viewpoint of \textit{attention shift}, in a similar manner to \cite{shi2020self}.
As visualized in Figure~\ref{fig: fcase}, the agent fails to judge decisive features and mistakenly attends to the task-irrelevant grassland, because the zigzag turn has never been encountered during training.
Catastrophic cumulative attention shifts would lead to problematic situations and cause significant performance degradation for practical RL agents.